\documentclass[10pt,conference]{IEEEtran}
\usepackage{cite}

\ifCLASSINFOpdf
   \usepackage[pdftex]{graphicx}
   \graphicspath{{figs/}}
   \DeclareGraphicsExtensions{.pdf,.jpeg,.png}
\else
   \usepackage[dvips]{graphicx}
   \graphicspath{{../figs/}}
   \DeclareGraphicsExtensions{.eps}
\fi

\usepackage[cmex10]{amsmath}
\usepackage{amsthm}

\usepackage{array}

\ifCLASSOPTIONcompsoc
  \usepackage[caption=false,font=normalsize,labelfont=sf,textfont=sf]{subfig}
\else
  \usepackage[caption=false,font=footnotesize]{subfig}
\fi

\usepackage{hyperref}
\usepackage{booktabs, multirow}

\newif\iffinal
\finaltrue

\iffinal
\else
\usepackage[switch]{lineno}
\fi

\newcommand\blfootnote[1]{%
  \begingroup
  \renewcommand\thefootnote{}\footnote{#1}%
  \addtocounter{footnote}{-1}%
  \endgroup
}

\begin{document}
\title{Towards robustness under occlusion for\\ face recognition}
\iffinal
\author{\IEEEauthorblockN{Tomas M. Borges}
\IEEEauthorblockA{Electrical Engineering Department \\
\textit{University of Brasilia}\\
Brasilia, Brazil \\
tomas.borges@aluno.unb.br}
\and
\IEEEauthorblockN{Teofilo E. de Campos}
\IEEEauthorblockA{CIC / IE / UnB, Brazil and\\
Vicon Motion Systems Ltd, UK\\
\url{https://cic.unb.br/~teodecampos}\\
t.decampos@oxfordalumni.org}
\and
\IEEEauthorblockN{Ricardo de  Queiroz}
\IEEEauthorblockA{Department of Computer Science \\
\textit{University of Brasilia}\\
Brasilia, Brazil \\
queiroz@ieee.org}
}

\else
  \author{}
  \linenumbers
\fi

\maketitle

\begin{abstract}
  In this paper we evaluate the effects of occlusions in the performance of a face recognition pipeline which uses a ResNet backbone.
The classifier was trained on a subset of the CelebA-HQ dataset containing 5,478 images from 307 classes, to achieve top-1 error rate of 17.91\%.
We designed 8 different occlusion masks which were applied to the input images.
This caused a significant drop in the classifier performance: its error rate for each mask became at least two times worse than before.
  In order to increase robustness under occlusions, we followed two approaches. The first is image inpainting using the pre-trained pluralistic image completion network. The second is Cutmix, a regularization strategy consisting of mixing training images and their labels using rectangular patches, making the classifier more robust against input corruptions.
  Both strategies revealed effective and interesting results were observed. In particular, the Cutmix approach makes the network more robust without requiring additional steps at the application time, though its training time is considerably longer.
  Our datasets containing the different occlusion masks as well as their inpainted counterparts are made publicly available to promote research on the field.\blfootnote{T.\ Borges is a PhD student supervised by R. de Queiroz and this piece of work started as an assignment done by T.\ Borges as part of the Computer Vision course given by T.\ de Campos at the University of Brasilia. We acknowledge the support of \href{http://cnpq.br/}{CNPq} through grant PQ 314154/2018-3 and \href{http://www.fap.df.gov.br}{FAPDF}.}
\end{abstract}


\IEEEpeerreviewmaketitle

\section{Introduction}
\label{sec:intro}
Modern face classifiers based on deep learning methods leverage large datasets to learn the essential features that make each face unique and are able to achieve high-performance on the recognition task, even outperforming human capabilities \cite{Taigman2014}.
However, in presence of occlusions, like masks, glasses, hair, or even food, face classifiers struggle \cite{Ekenel2009, Zeng2020}.

Face detection is the first step for recognition, the general case approaches consist of using rigid templates, \emph{e.g.}, the Viola-Jones face detector \cite{Viola2001}; deformable part models (DPM) \cite{Yan2014}; and deep learning models, like the Sigle-Shot Mutibox Detector (SSD) \cite{Liu2016}, for example.
When extremely large datasets \cite{Yang2016} are used to train deep learning methods, those are able to detect occluded faces \cite{Zhang2020}, still there are other approaches specialized in detection of faces under occlusion.
Those methods are commonly tested over the masked faces in the wild (MAFA) dataset \cite{Ge2017} and employ strategies such as locating visible facial segments to estimate the face \cite{Yang2018}, fusing detection results obtained from face sub-regions \cite{Mahbub2016}, or considering occluded faces as a particular single-class object detection problem \cite{Opitz2016}.

Usually the approaches to solve occluded face recognition (OFR) can be divided into three categories \cite{Zeng2020}:
\begin{enumerate}
    \item In occlusion-robust feature extraction approaches, the aim is to extract features that are less affected by occlusions, while preserving the discriminative capability.
    Feature extraction can be performed by engineered-based methods, such as local binary patterns (LBP) \cite{Ahonen2004}, and scale-invariant feature transform (SIFT) descriptors \cite{Lowe1999}, or by learning-based methods, in which deep convolution neural networks (CNN) approaches stand out by performing the training with augmented occluded faces and
    reducing the spatial support for a more
    local feature extraction \cite{Osherov2017}.  
    \item In occlusion-aware face recognition, the idea is to exclude the occlusion area, such that only part of the face is used for the classification task \cite{Wan2017,Song2019}.
    \item Finally, there are occlusion-recovery-based approaches that try to solve the occlusion problem in the image space by completing the occluded area.
    Again, deep learning techniques stand out either when explicitly addressing de-occlusion of faces \cite{Zhao2018,Li20}, or when used for blind inpainting through generative models employing variational autoencoders (VAE) \cite{Kingma2013}, generative adversarial networks (GAN) \cite{Goodfellow2014, He2019, Ge2020}, and partial convolutions \cite{Liu2018}.
\end{enumerate}

We used a dataset formed by images containing only aligned faces, such that face detection was not needed.
Our focus is thus on the classification part of an OFR system.
We employed data-driven approaches to present an occlusion-robust classifier leveraging robust feature extraction, and a recovery-based strategy for image completion for the occluded face classification.
The tested occlusions were artificially added to the images utilizing binary masks and filling-in with a mid grey color.

Besides the benchmarks achieved in this paper, our contribution also comprises the resulting publicly available datasets with different kinds of synthetic occlusions and their inpainted counterparts.
To the best of our knowledge, this group of datasets is quite unique in the literature and surely can be helpful to facilitate and promote research on the field.

\section{Datasets}
\label{sec:datasets}
\subsection{CelebA-HQ}
\label{sec:celeba-hq}
\begin{figure*}[ht]
    \centering
    \includegraphics[width=\textwidth]{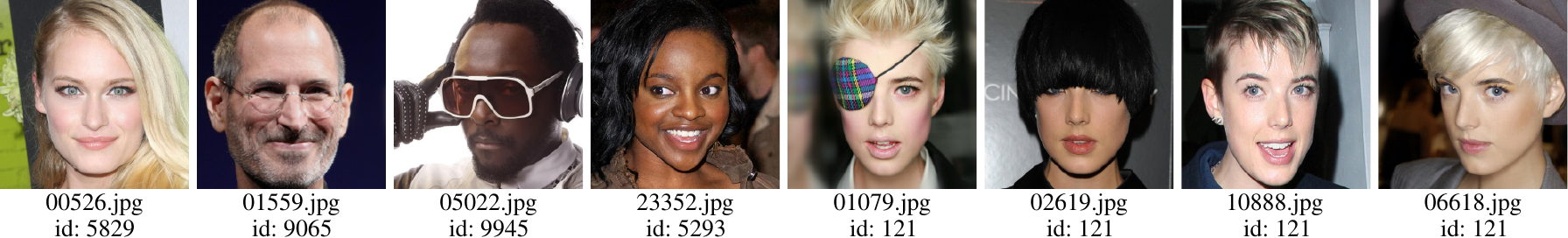}
    \caption{Examples of the CelebA-HQ dataset.}
    \label{fig:celeba_hq}
\end{figure*}

To avoid the need for face detection and to concentrate our efforts in the classification task, we choose the CelebFaces Attributes Dataset (CelebA) \cite{Liu2015}.
More specifically, we utilized the aligned and cropped version pre-processed for super-resolution at 1024-by-1024 pixels, also known as the CelebA-HQ dataset comprising of 30,000 images \cite{Karras2018, CelebA-HQ}.
In the CelebA-HQ dataset, the number of images is not equally divided among classes, so we worked only with a subset of it, in which each class (or person) appears, at least, in 15 images.
Thus, the dataset was reduced from 30,000 to 5,478 images of center-aligned faces from 307 celebrities.

The dataset was randomly divided into three fixed sets: a training set (3,943 images), with at least 10 images of each class; a validation set (921 images), with 3 images of each class; and a test set (614 images), with 2 images of each class.
Figure \ref{fig:celeba_hq} depicts examples of the dataset.
Looking through the images we noticed some characteristics worth mentioning:
\begin{enumerate}[i)]
    \item Although most of the pictures are front-facing, there is some pose variation;
    \item There are some faces with natural occlusions (hair, hats, glasses, hands, microphone);
    \item There are some images with distorted background;
    \item The dataset is not well-balanced, blond white women are the majority;
    \item Celebrities tend to change a lot their appearance, making the classification task harder.
\end{enumerate}

\subsection{Synthetic occlusions}
\label{sec:synth_occlusions}
\begin{figure}[ht]
    \centering
    \includegraphics[width=\columnwidth]{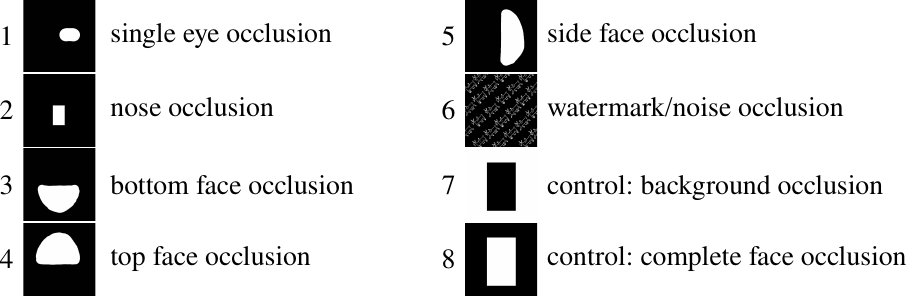}
    \caption{Binary masks used to simulate occlusions (white pixels).}
    \label{fig:masks}
\end{figure}

To simulate occlusions we defined 8 binary masks to account for different parts of the face, as depicted in Figure \ref{fig:masks} and discussed below.

\begin{itemize}
\item Masks 1 and 2 represent minor occlusions, removing only one eye (left or right eyes are occluded randomly) or the nose, respectively.
\item Mask 3 simulates occlusions caused by surgical masks, which, nowadays, became a quite common occlusion artifact.
\item Mask 4 occludes both eyes and most of the hair, which is a typical occlusion used to anonymize identity.
\item Mask 5 occludes one entire side of the face to simulate occlusion by pose variation (the side of the occlusion is randomly chosen).
\item Mask 6 simulates occlusion by watermarking or by some degradation in the picture.
\item Masks 7 and 8 are the control group, with the former isolating the face and occluding only the background, where minimal interference in detection is expected, while the latter completely occludes the face and correct matches are only expected by chance.
\end{itemize}

\begin{figure}[ht]
    \centering
    \includegraphics[width=\columnwidth]{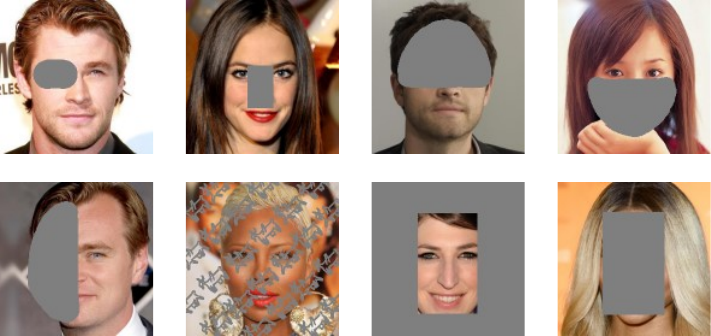}
    \caption{Examples of the eight resulting test datasets with occlusions.}
    \label{fig:datasets_w_occlusions}
\end{figure}

Each of the occlusions were applied only to the test dataset, thus resulting in eight new test sets.
The occluded areas were filled-in with a mid grey color.
Figure \ref{fig:datasets_w_occlusions} depicts one example of each such datasets.

\subsection{Inpainting}
\label{sec:inpainting}
Inpainting, or image completion, is the process of reconstructing the missing parts of an image.
The inpainted region should have semantic cohesion with the overall scene, continuity around the edges, and fill-in visually realistic content.
It is quite subjective since there is no prior knowledge about missing parts, and many completions may appear correct but may modify cues that may be very relevant for a person's identity.

There are many approaches for solving inpainting problems, as mentioned in Section \ref{sec:intro}.
For this paper we choose to use the pluralistic image completion (PICNet) \cite{Zheng2019}.
It is a probabilistically principled framework, which during training, uses the original image $I_{gt}=\{I_m,~I_c\}$, a degraded version $I_m$ (the masked partial image), and $I_c$ its complement partial image, to generate inpainted versions to $I_m$ by sampling from $p(I_c|I_m)$.

\begin{figure*}[hbt]
    \centering
    \includegraphics[width=.7\textwidth]{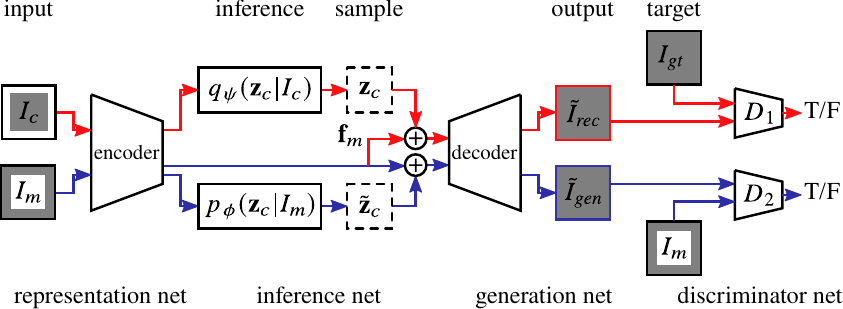}
    \caption{Simplified diagram of the architecture of the PICNet \cite{Zheng2019}.}
    \label{fig:picnet}
\end{figure*}

The PICNet is implemented with two parallel paths, as illustrated in Figure \ref{fig:picnet}.
The upper path is the reconstructive pipeline (red lines), which utilizes $I_c$ to infer the importance function $q_{\psi}(\textbf{z}_c|I_c)$ during training.
Using the sampled latent vector $\textbf{z}_c$ and the conditional feature $\textbf{f}_m$, which encodes the information of the visible regions from $I_m$, there is enough information to train the decoder to reconstruct the original image $I_g$.

The lower path is the generative pipeline (blue lines), which only uses $I_m$ to infer the conditional prior $p_{\phi}(\textbf{z}_c|I_m)$ of the holes.
In this case the sampled latent vector $\tilde{\textbf{z}}_c$ is not accurate enough to reconstruct the whole image, thus the decoder targets to reconstruct only the visible regions $I_m$, using $\textbf{f}_m$.
Both paths are supported by GANs, to ensure that the synthesized data fit in with the training set distribution, leading to higher-quality images.
Additionally, there is a short+long term attention layer that exploits distant relations among decoder and encoder features, improving appearance consistency.
Both representation and generation networks share the exact same weights, and testing is performed only in the generative path, with $\tilde{I}_{\emph{gen}}$ the inpainted output.

\begin{figure}[bt]
    \centering
    \includegraphics[width=\columnwidth]{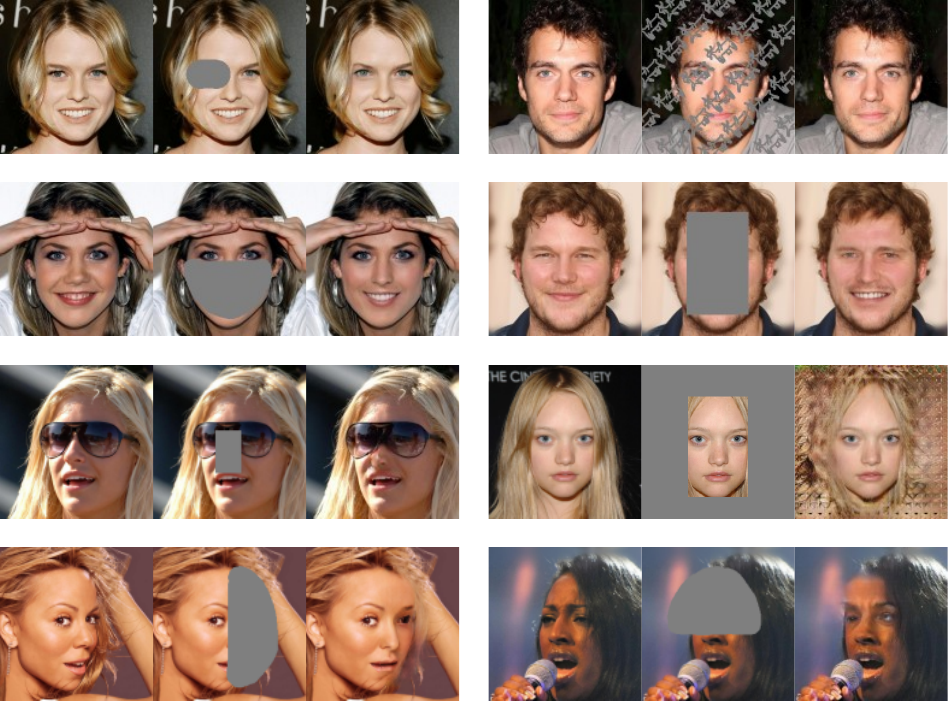}
    \caption{Examples of inpainting using PICNet (original-masked-inpainted). The reconstruction of images in the first row is quite faithful to the original images. Images in the second row have a high-quality reconstruction, although quite different from its original counterpart. The last two rows show
    cases that can be considered as failure, as PICNet introduced unexpected artefacts and distorted faces.
    }
    \label{fig:inpainting_datasets}
\end{figure}

In our evaluations, we used a model trained on the CelebA-HQ dataset using random occlusion masks \cite{Zheng2019}.
Figure \ref{fig:inpainting_datasets} shows some examples of the inpaintings produced by PICNet.

Although the PICNet is able to generate multiple and diverse plausible solutions for image completion, the inpainted datasets were created using only one solution for each image.
\section{Face classification}
\label{sec:face_classification}
\subsection{Regular Classifier}
\label{sec:regular_classifier}

For the face classification task we chose to use a ResNet \cite{He2016} architecture.
Since the chosen dataset did not provide many images from each class (15 up to 28 images for each person), data augmentation was performed.
We performed data augmentation by applying random variations of the input data for the training set images, such that the images became different, but their meaning is unchanged. These include flipping, rotation, zooming, warping, and lighting transformations.

We started training our classifier with pre-trained ResNets from \texttt{fastai}  \cite{Howard2020}, trained on ImageNet \cite{ImageNet}.
Pre-trained weights make for a better initialization than random weights, since the layers are already suited to extract meaningful features for image classification.
A cross entropy loss function with flattened input was used to compare predictions and targets.
To define the size of the network, we trained ResNets with 18, 34, 50, and 101 layers.
The training for all those networks was performed using 1 frozen epoch and 25 unfrozen ones.
The learning rate was scheduled using cosine annealing and momentum \cite{Smith2019}, with its maximum at $\lambda=5\cdot10^{-3}$, found using the technique from Leslie Smith \cite{Smith2017}.
The results for the validation and test sets are shown in Table \ref{tab:resnets}.
For the regular classifier and for the experiments in the remaining of this paper, we chose to use the ResNet101, since it provided the lowest error rate in the test set. Since the purpose of the experiments in this section was to establish the baseline method that uses non-occluded faces and to decide the architecture to be used for comparison, it was fair to use the test set in this decision. In the remainder of this paper, any hyper-parameter tuning was done based on the validation set.

\begin{table}[htb]
    \centering
    \caption{Baselines error rates for different ResNet depths considering the test set without occlusions.}
    \label{tab:resnets}
    \begin{tabular}{@{}lccc@{}}
        \toprule
        \multirow{2}{*}{Architecture} & \multirow{2}{*}{Validation error} & \multicolumn{2}{c}{Test error}     \\ \cmidrule(l){3-4} 
                            &                                   & Top-1            & Top-5           \\ \midrule
        ResNet18          & 25.52\%                           & 23.94\%          & 11.07\%         \\
        ResNet34          & 22.37\%                           & 23.62\%          & 10.10\%         \\
        ResNet50          & \textbf{19.00\%}                  & 20.52\%          & 8.47\%          \\
        ResNet101         & 19.98\%                           & \textbf{17.91\%} & \textbf{7.00\%} \\ \bottomrule
    \end{tabular}
\end{table}
\subsection{Robust Classifier}
\label{sec:robust_classifier}
\begin{figure}[htb]
    \centering
    \includegraphics[width=\columnwidth]{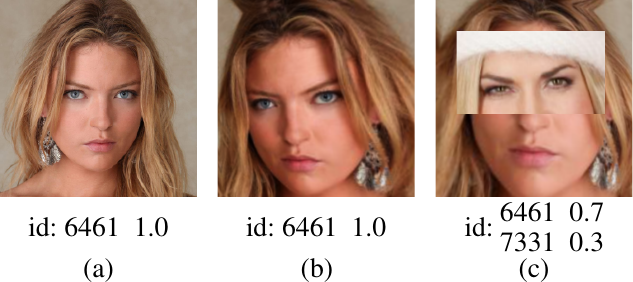}
    \caption{Examples of data augmentation. (a) Original image. (b) Output of data augmentation transforms. (c) Cutmix, notice the label in this case is a mixture of the two images. }
    \label{fig:data_aug}
\end{figure}

To train the robust classifier we added the Cutmix \cite{Yun2019} approach as a regularization strategy.
The idea of this method is to cut and paste rectangle patches among training images, where the labels are also mixed proportionally to the area of the patches.
This approach prevents the CNN from focusing too much on a small set of intermediate activations or on a small region of input images, which improves generalization and localization by forcing the model to attend to the entire object.
Also, since the patches are formed by other images from the dataset and the ground truth labels are also mixed, this approach further enhances localization ability by requiring the model to identify the object from a partial view.
Therefore, Cutmix adds exactly the kind of robustness desired for classifying images under occlusion.
Figure \ref{fig:data_aug} illustrates the data augmentation strategies applied in the training images.

Training with Cutmix is considerably slower, since it becomes harder for the model to learn from the mixed examples. This is the case for any method that deals with noisy labels~\cite{cordeiro2020tutorial}.
The training was performed the same way as before, using the transfer learning procedure and learning rate schedule, but this time we had to train the model for 70 unfrozen epochs to reach a validation error of \textbf{19.00\%}. The obtained model reaches top-1 and top-5 test errors of \textbf{17.42\%} and \textbf{7.98\%}, respectively. As expected, this is a little bit better than the original method, but the obtained network should have higher generalization power.
\section{Performance assessment and analysis}
\label{seq:analysis}
\subsection{Experiment setup}
\label{sec:setup}
In order to check how the occlusions affect the classification task we use as baselines the results obtained in Section \ref{sec:face_classification} for the top-1 error over the test set without occlusions: 17.91\% and 17.42\%, respectively for the regular and for the robust classifiers.
Then, we check, for each of the test sets with occlusions, how the classifiers perform.
Finally, we compare these results with the classification error rates for the inpainted datasets.


\subsection{Results}
\label{sec:results}
\begin{figure}[ht]
    \centering
    \includegraphics[width=\columnwidth]{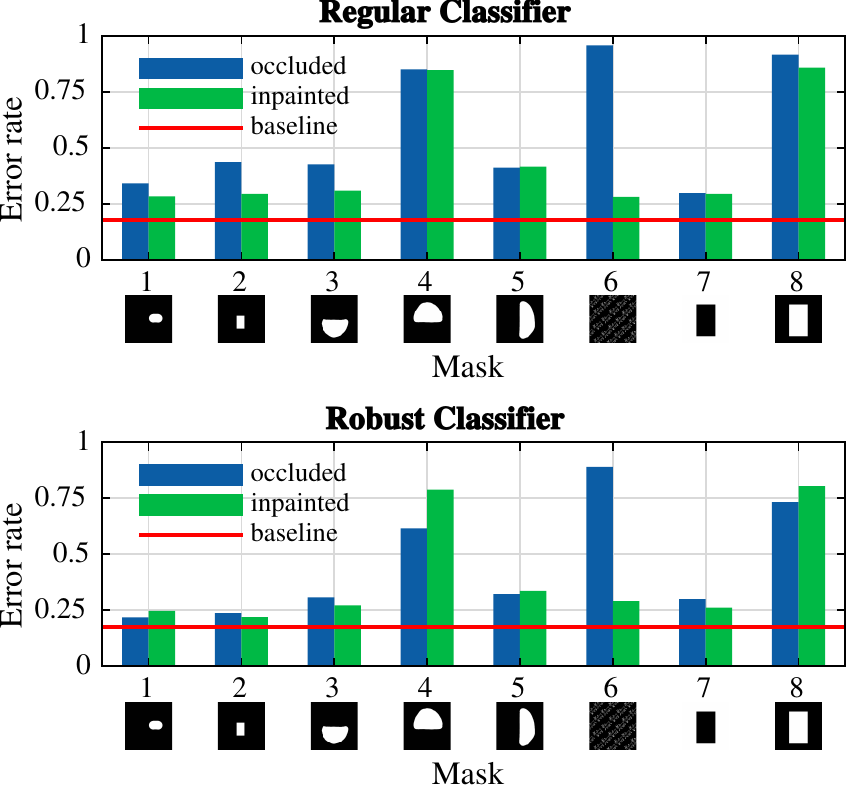}
    \caption{Performance of the two classifiers over different occlusions and inpaintings. Top: ResNet101 regular training. Bottom: ResNet101 with Cutmix.}
    \label{fig:results}
\end{figure}

After adding the synthetic occlusions, we noticed that the error rate increased for both classifiers in all tested occlusion masks, the dark blue bars in Figure \ref{fig:results}.
As for the occlusion-recovery-based approach with the inpainted images (green bars), we can see that it was not always helpful and its behavior is heavily dependent on the classifier's robustness.

From Figure \ref{fig:results}-top, we observe that the regular classifier was severelly affected by the occluded images, having its error rate doubling for all cases, except for mask 7.
The inpainted images did improve this classifier's performance, specially in the presence of small occlusions, and for the watermark case (mask 6).

The robust classifier in Figure \ref{fig:results}-bottom was indeed better at ignoring the occlusions, having lower error rates for all cases.
One interesting thing is that the inpainting did worsen the robust classifier's performance for some masks (1, 4, and 5), this makes sense when we look at some of the inpainted images and see that, in some cases, the completions completely changes subject's appearance.

\begin{figure*}[t]
    \centering
    \includegraphics[trim={0 0 0 0},clip,width=.96\textwidth]{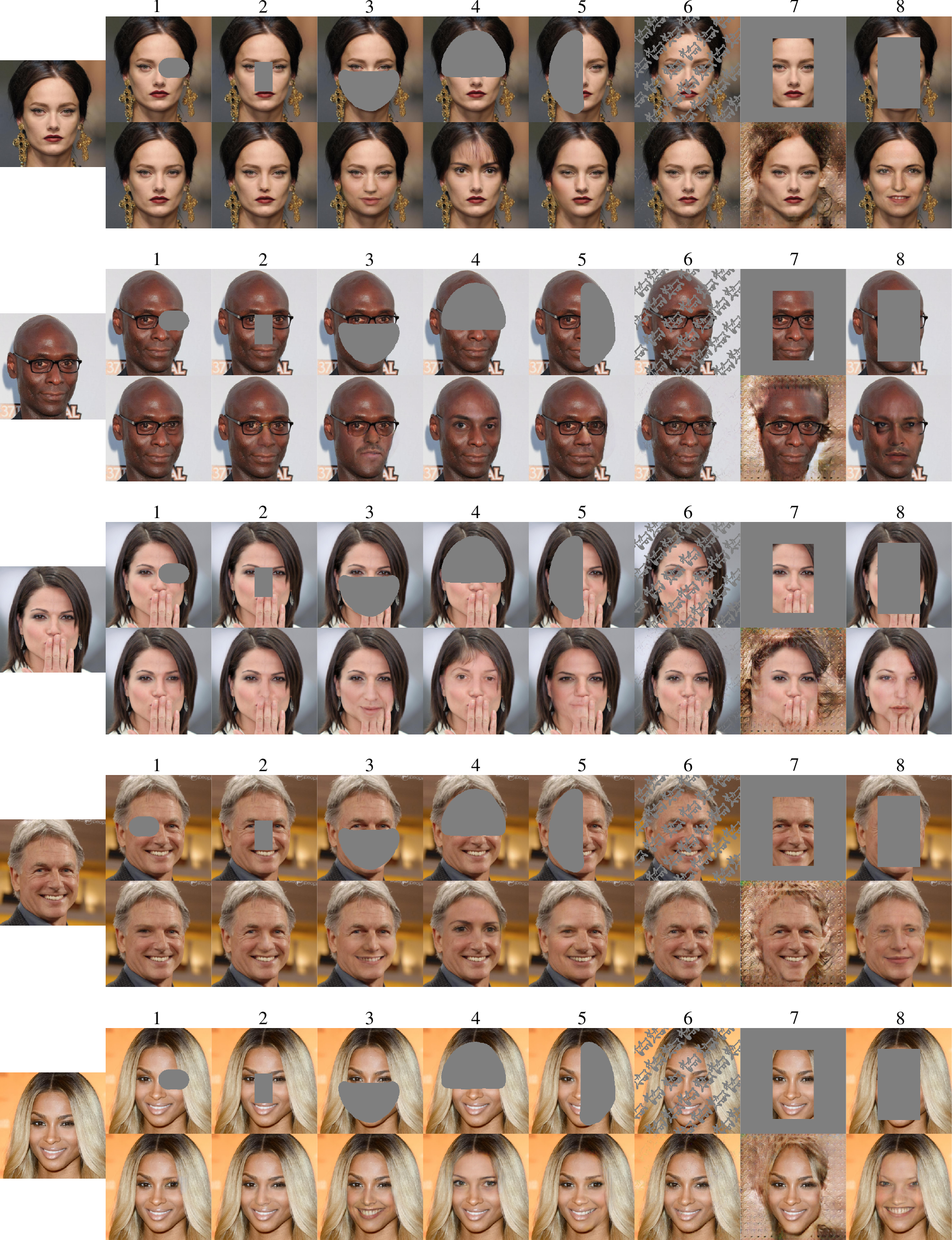}
    \caption{Example of inpainting results for all the masks.}
    \label{fig:inpaintings}
\end{figure*}

The inpainting results for masks 1 and 5 were a little disappointing since the short+long term attention layer of the PICNet should supposedly ``copy'' the symmetric information, such that, in those cases, a lower error rate was expected.
Results for mask 4 are expectedly high since this kind of occlusion is indeed made to hide identity (for anonymization purposes), and inpainting in this case could not recover the true features of the faces.
The best performance gains for the inpainting technique happened with the heavy watermark mask (6). This kind of occlusion was the worst for both classifiers and inpainting brought the error rate near the non-occluded baseline.
A surprising result was that of mask 7, which we believed would not have caused any harm on either classifier because it only masks out the areas of the images which are near the borders. However, that mask degraded both classifiers, indicating that ResNet (even if trained with CutMix) clearly uses non-facial cues for person recognition. Furthermore faces in this dataset are not aligned and relevant identification information lie near the border of some images. 
Results in mask 8 are as expected, both classifiers failed to deliver any sensible classification with or without inpainting, since most of the subjects' faces are not visible with that mask.
In Figure \ref{fig:inpaintings} we illustrate some examples of the inpainting results for all the masks.
\section{Conclusions}
In this paper we studied two ways to deal with the problem of face recognition under occlusions (OFR) and 
created 16 modified test sets which can be used for OFR problems, as well as for detection of deep fakes.
In our results, training a robust classifier was the strategy that generated better results, but for almost every case, inpainting was helpful for improving the accuracy of the classifier.

For future works, we propose to investigate how much of the techniques presented here could be transferred for situations in which occlusion masks are not available and have to be inferred by semantic segmentation.
Furthermore, it is well known the robustness of face recognition systems is improved by registering faces by aligning their features (e.g.\ eyes and nose tip). However, our results with mask 7 (which only occludes pixels near the image border) may indicate that some peripheral features (such as hair and neck) actually have a relatively high importance on face recognition. It will be interesting to investigate that systematically, using facial feature detection and explainable AI techniques. 

Finally, it will be interesting to make use of more than one output of the pluralistic image completion approach to (i) check if there is considerably variability in the results and (ii) use an ensemble strategy to combine results of multiple inpainted versions of each image.


\bibliographystyle{IEEEtran}

\bibliography{refs}

\end{document}